\documentclass[letterpaper, 10 pt, conference]{ieeeconf}  

\IEEEoverridecommandlockouts                              

\usepackage[T1]{fontenc}
\usepackage{amssymb}
\usepackage{amsmath}
\usepackage{babel}
\usepackage{graphicx,import}

\usepackage{etoolbox}
\usepackage{wrapfig}
\usepackage{microtype}
\usepackage{booktabs} 
\usepackage{siunitx}
\usepackage[table]{xcolor}
\usepackage{comment}
\usepackage[breaklinks, citecolor=black]{hyperref}
\usepackage{tikz}
\usepackage{subfig}
\usepackage{graphicx}
\usepackage{float}

\newcommand\gabrisays[1]{\textcolor[rgb]{0.7, 0.2, 0}{Gabri: #1}}

\definecolor{Gray}{gray}{0.9}
\newcolumntype{g}{>{\columncolor{Gray}}c}

\makeatletter
\patchcmd{\@makecaption}
  {\scshape}
  {}
  {}
  {}
\makeatletter
\patchcmd{\@makecaption}
  {\\}
  {.\ }
  {}
  {}
\makeatother

\graphicspath{ {./images/} }

\title{\LARGE \bf
Nonlinear Model Identification and Observer Design for Thrust Estimation of Small-scale Turbojet Engines
}

\author{Affaf Junaid Ahamad Momin$^{1,2}$, Gabriele Nava$^{1}$, Giuseppe L'Erario$^{1,3}$, Hosameldin Awadalla Omer Mohamed$^{1,2}$, \\ Fabio Bergonti$^{1,3}$, Punith Reddy Vanteddu$^{1}$, Francesco Braghin$^{2}$, and Daniele Pucci$^{1,3}$

\thanks{$^{1}$ Artificial and Mechanical Intelligence, Istituto Italiano di Tecnologia, Genova, Italy {\tt\small firstname.surname@iit.it}}%
\thanks{$^{2}$ Department of Mechanical Engineering, Politecnico di Milano, Milan, Italy {\tt\small francesco.braghin@polimi.it} {\tt\small affafjunaid.momin@mail.polimi.it}}%
\thanks{$^{3}$ School of Computer Science, Univ. of Manchester, Manchester, U.K.}
}

\begin{document}

\maketitle
\thispagestyle{empty}
\pagestyle{empty}

\begin{abstract}

Jet-powered vertical takeoff and landing (VTOL) drones require precise thrust estimation to ensure adequate stability margins and robust maneuvering. Small-scale turbojets have become good candidates for powering \emph{heavy} aerial drones. However, due to limited instrumentation available in these turbojets, estimating the precise thrust using classical techniques is not straightforward. In this paper, we present a methodology to accurately estimate the online thrust for the small-scale turbojets used on the iRonCub - an aerial humanoid robot. We use a grey-box method to capture the turbojet system dynamics with a nonlinear state-space model based on the data acquired from a custom engine test bench. This model is then used to design an extended Kalman filter that estimates the turbojet thrust only from the angular speed measurements. We exploited the parameter estimation algorithm to ensure that the EKF gives smooth and accurate estimates even at engine failures. The designed EKF was validated on the test bench where the mean absolute error in estimated thrust was found to be within 2\% of rated peak thrust.

\end{abstract}



\section{INTRODUCTION}
\label{sec:introduction}

Unmanned Aerial Vehicles (UAVs) feature different types of propulsion devices ranging from electric-driven propellers to jet engines. Many modern UAVs employ jet engines due to their high energy and power densities~\cite{pucci2017momentum,jetquad}. The iRonCub - a VTOL humanoid robot propelled by four small-scale turbojets - is a relevant example~\cite{nava2018position,hosam_thrust}. In vehicles like these, accurate thrust estimation is crucial in order to guarantee adequate stability margins and perform dynamic maneuvers. Since direct measurement of in-flight engine thrust with force sensors is often not feasible due to on-board complexities, the online thrust is usually calculated from engine state measurements and engine models. However, most commercial small-scale turbojets do not feature the comprehensive instrumentation that is readily available in full-sized jet engines. Because of limitations in engine state measurements, estimating the thrust accurately becomes a challenge. In this paper, we present a methodology to obtain accurate and robust real-time thrust estimates for small-scale turbojets using only the live angular speed measurements.

Several physics-based and empirical models of turbojet engines have been explored in literature~\cite{asgari_survey_paper}. While physics-based models are built exploiting the governing principles of turbojet~\cite{Klein_gasturb,going_P200,single_spool_turbojet_dynamic}; data-driven models are constructed only from experimental data often disregarding intrinsic system characteristics~\cite{chiras_ffw_neural,asgari_single_shaft,ruano_spey_dynamics}. Grey-box or hybrid models that combine the two approaches have also been explored in~\cite{catana_gt_instrumentation,turbojet_acceleration_mode,mohammadi_montazeri}. Basically, all types of engine models use a given set of measurements like the pressure ratio, fuel flow rate, etc, to calculate unknown engine states such as the thrust.\looseness=-1

Of particular relevance to small-scale turbojets is the work of L'erario et al~\cite{L'Erario2020}. They proposed a second-order nonlinear state-space model to express the thrust dynamics using a data-driven approach. Their experiment involved applying input control signals to the turbojet and measuring its thrust response with a high-bandwidth force sensor in a custom test bench. However, engine state measurements were not used in their study. Similar work on a small-scale turbojet engine was carried out by Jiali et al~\cite{Jiali2015}.

Even though turbojet models are abundant in existing literature, pragmatic approaches to estimate the accurate thrust considering measurement and model uncertainties are quite sparse. A data-driven Hammerstein-Wiener model for a UAV turbojet was proposed in the work of Villarreal-Valderrama et al~\cite{direct_thrust_hamm_wiener}. They used a linear Kalman Filter to estimate thrust only from angular speed measurements. However, the accuracy of the estimator for fast-dynamic maneuvers and its robustness to engine failures was not investigated.

In this paper, we use an identification technique called SINDy~\cite{Brunton3932} to construct a nonlinear dynamic model for the turbojet angular speed. We generate key insights into system characteristics from our experiments on a custom engine test bench, and incorporate those into the data-driven model. This model is then used to design a robust EKF that estimates the accurate online engine thrust only from angular speed measurements, and exploits the parameter estimation algorithm to maintain the smoothness and accuracy of thrust estimates even during events of engine failure.

The paper is organized as follows. Sec.~\ref{sec:background} presents the notations used, the fundamentals of turbojet physics, and some technical details of the small-scale turbojet engines used in this study. Sec.~\ref{sec:identification} contains the description of the experimental setup, the grey-box procedure used for identifying the nonlinear turbojet models and validation results. Then, Sec.~\ref{sec:ekf_thrust} deals with the design of the extended Kalman Filter to estimate the online turbojet thrust. Sec.~\ref{sec:ekf_validation} discusses the performance and validation results of the designed EKF. Finally, Sec.~\ref{sec:conclusions} presents the conclusions of this study and prospects of future work.

\section{BACKGROUND}
\label{sec:background}

\subsection{Notation} 
\begin{itemize}
    \item  $\mathbf{x} \in \mathbb{R}^n$ - state vector of a system
    \item  $\mathbf{u} \in \mathbb{R}^p$ - input vector
    \item  $\mathbf{y} \in \mathbb{R}^m$ - output vector or measurement vector
    \item  $\mathbf{Q} \in \mathbb{R}^{n \times n}$ - process noise covariance
    \item  $\mathbf{R} \in \mathbb{R}^{m \times m}$ - measurement noise covariance
    \item  $\mathbf{P} \in \mathbb{R}^{n \times n}$ - state estimation error covariance
    \item  $T$ - thrust produced by the turbojet [$\mathrm{N}$]
    \item  $\omega$ - turbojet angular speed [$\mathrm{kRPM}$]
    \item  $u \in [0,100]$ - input signal to the turbojet
    \item  $m_a$ - mass flow rate of air through the turbojet [$\mathrm{kg/s}$]
    \item  $V_a$ - volumetric flow rate of air [$\mathrm{m^3/s}$]
    \item  $v_e$ - velocity of exhaust gas [$\mathrm{m/s}$]
    \item  $\rho_{a}$ - air density [$\mathrm{kg/m^3}$]
\end{itemize}

\subsection{Overview of Turbojet Physics}

\begin{figure}[t]
    \centering
    \includegraphics[scale=0.09]{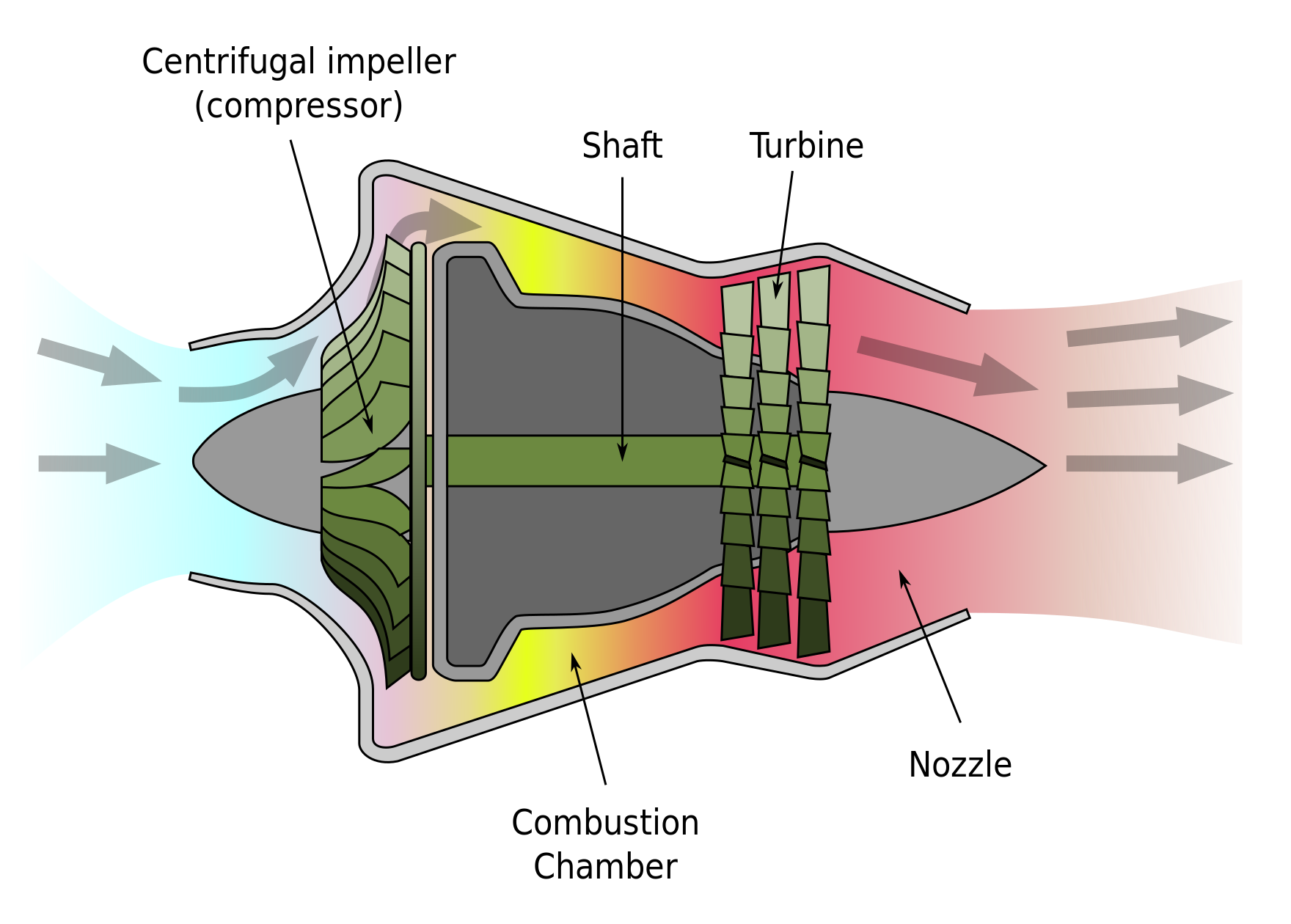}
    \caption{Turbojet with Centrifugal Compressor. Source~\cite{turbojet_scheme_pic}}
    \label{fig:1}
\end{figure}

Let us recall the basic principles of the turbojet engine~\cite{flack_jet_basics},~\cite{cumpsty_jet_basics}. A turbojet is a rotary device that converts heat energy released by burning fuel into thrust or mechanical energy. It achieves this by following the Brayton cycle~\cite{wu2007thermodynamics}. Fig.~\ref{fig:1} shows a schematic diagram of a turbojet with centrifugal compressor.

Assuming a free-stream air inlet velocity of zero and neglecting the mass flow rate of fuel, the thrust produced by a subsonic turbojet is given by the following expression~\cite{flack_jet_basics}:
\begin{align} 
    T = \dot{m}_{a} v_{e} = \rho_{a} \dot{V}_{a} v_{e} \label{eq:6}.
\end{align}

For a given turbojet compressor operating at a constant angular speed, the volumetric flow rate of air $\dot{V}_a$ depends on the angular speed~\cite{Brown1991FanLT}. The velocity of exhaust $v_e$ depends on the angular speed and the fuel flow rate~\cite{flack_jet_basics}. Hence, from Eq.~\eqref{eq:6}, the thrust produced depends on: the angular speed of the turbojet, the fuel flow rate, and the air density.

\subsection{Turbojet specifications and architecture}
In this study, two different models of  small-scale turbojet engines were used: \textit{(a)} JetCat P160-RXi-B~\cite{P160_ref}, and \textit{(b)} JetCat P220-RXi~\cite{P220_ref}.
\begin{figure}[t]
      \centering
      \includegraphics[scale=0.13]{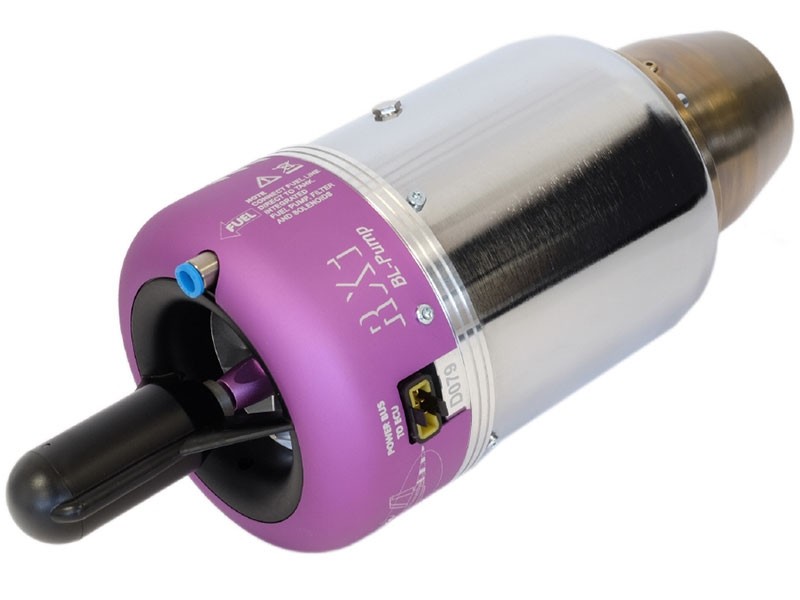} 
      \includegraphics[scale=0.31]{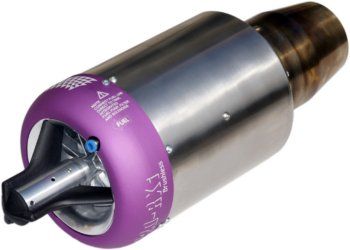}
      \caption{JetCat Turbojet Engines P160-RXi-B \textit{(left)} and P220-RXi \textit{(right)}.}
      \label{fig:2}
\end{figure}
\begin{figure}[t]
    \centering
    \includegraphics[scale=0.17]{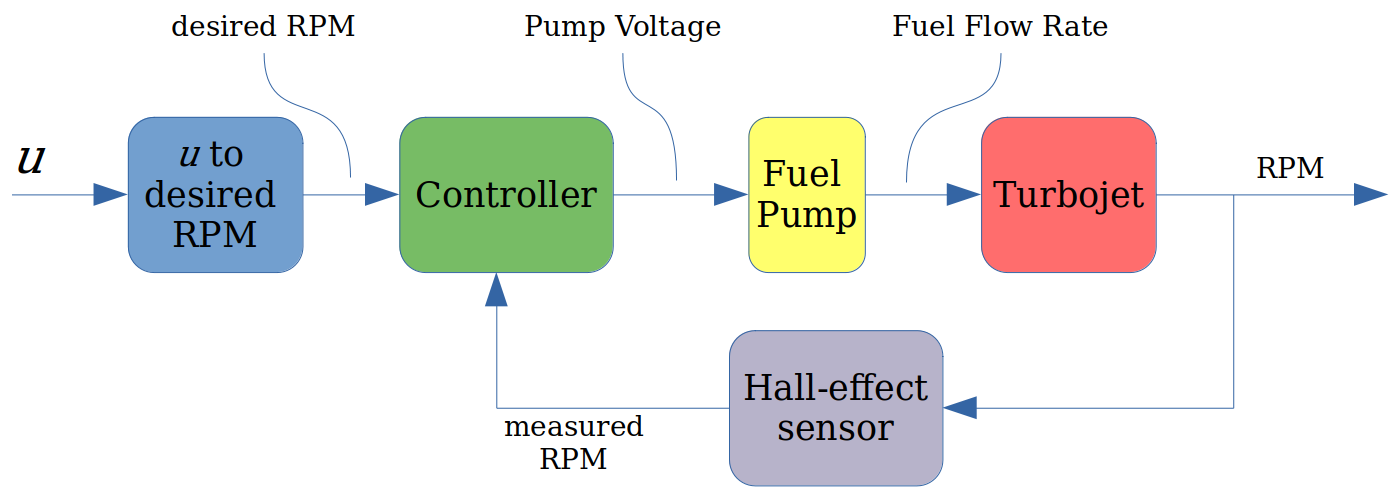}
    \caption{Internal control scheme of the turbojet}
    \label{fig:turbojet-control-scheme}
\end{figure}

\begin{table}[t]
\caption{Turbojet specifications.}
\label{tab:turbines_specs}
    \centering
    \begin{tabular}{c|c|c}
    \toprule
    \textbf{Parameter} & \textbf{JetCat P160} & \textbf{JetCat P220}  \\
    \midrule
    {Idle angular speed}              & $33000$ RPM & $35000$ RPM \\
    {Max angular speed}               & $123000$ RPM & $117000$ RPM \\
    {Thrust at idle speed}    & $7$ N & $9$ N \\
    {Thrust at max speed}     & $158$ N & $220$ N\\
    \bottomrule
    \end{tabular}
\end{table}

Both engines have a single-stage centrifugal compressor, a single-stage axial turbine and an internal fuel pump. Table~\ref{tab:turbines_specs} lists the turbojet specifications reported by the manufacturer at standard atmospheric conditions.

The turbojet is controlled by a digital input control signal $u\in[0,100]$. This input signal is mapped to a desired turbine angular speed depending on the control mode we choose. The turbojet has two steady-state control modes, namely:
\begin{itemize}
    \item \textit{Speed-proportional control} where the desired angular speed is set linearly proportional to $u$;
    \item \textit{Thrust-proportional control} where the desired angular speed is set so that the thrust is linearly proportional to~$u$. \looseness=-1 
\end{itemize}
In both of these modes, $u=0$ corresponds to idle speed and $u=100$ corresponds to maximum speed. We used the Thrust-proportional control mode for this study.

A simplified schematic of the internal turbojet control architecture is shown in Fig.~\ref{fig:turbojet-control-scheme}. The Engine Control Unit (ECU) takes $u$ as the input signal and maps it to a desired value of angular speed, $\omega_{des}$. Then, a feedback control loop alters the fuel flow rate by operating on the pump voltage to attain the desired angular speed.
\section{SYSTEM IDENTIFICATION OF TURBOJET}
\label{sec:identification}

\subsection{Experimental setup}

\begin{figure}[t]
    \centering
    \includegraphics[scale=0.17]{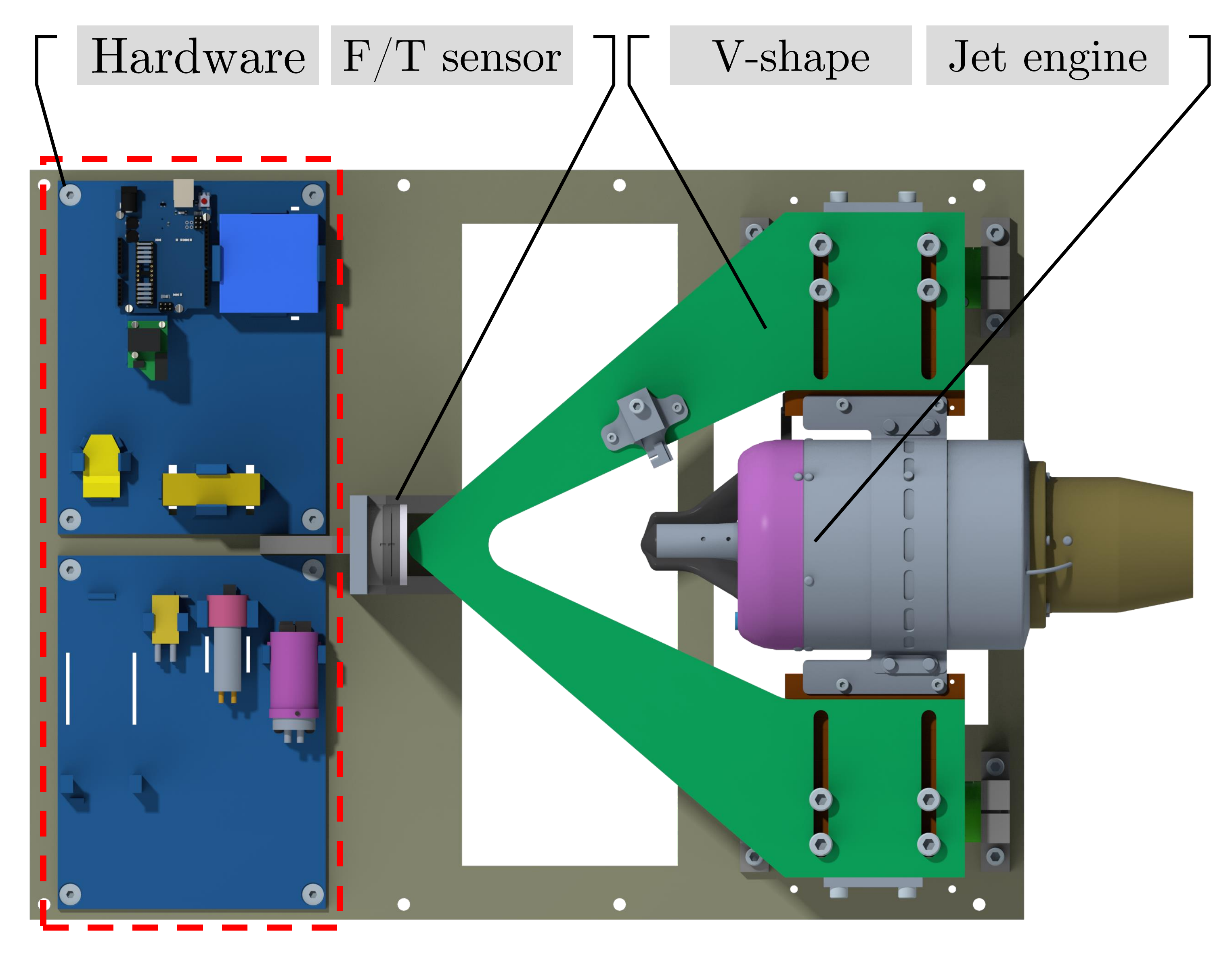}
    \caption{Test bench hardware setup.}
    \label{fig:setup}
\end{figure}

The experimental test bench setup for this study is the one used by L'Erario et al.~\cite{L'Erario2020}. The test bench setup scheme is presented in Fig.~\ref{fig:setup}.
The turbojet is rigidly mounted on a V-shape that moves on linear bearings. Thrust is measured by a precision 6-axis Force-Torque (F/T) sensor mounted in the setup when the V-shape comes into contact with it. The engine states are measured by internal sensors and streamed into the data logger. The entire architecture runs at a sampling rate of 100 Hz. Table \ref{table:measure} contains details on the measured quantities in the setup. Note that the angular speed measurements are quantized in steps of $100$ RPM.

\begin{table}[t]
\caption{Measured Quantities}
\label{table:measure}
    \centering
    \begin{tabular}{c|c|c|c}
    \toprule
    \textbf{Quantity} & \textbf{Sensor} & \textbf{Units} & \textbf{Resolution}  \\
    \midrule
    {Thrust}                    & 6-axis F/T & N & 0.25 N \\
    {Rotor Angular Speed}       & Hall effect & RPM & $100$ RPM \\
    {Pump Voltage}              & Voltmeter & V & $0.01$ V  \\
    {Exhaust Gas Temperature}     & Thermocouple & $^oC$ & $1 ^oC$ \\
    \bottomrule
    \end{tabular}
\end{table}

\subsection{Model structure identification}

\begin{figure}[t]
\centering
 \includegraphics[width=0.48\textwidth]{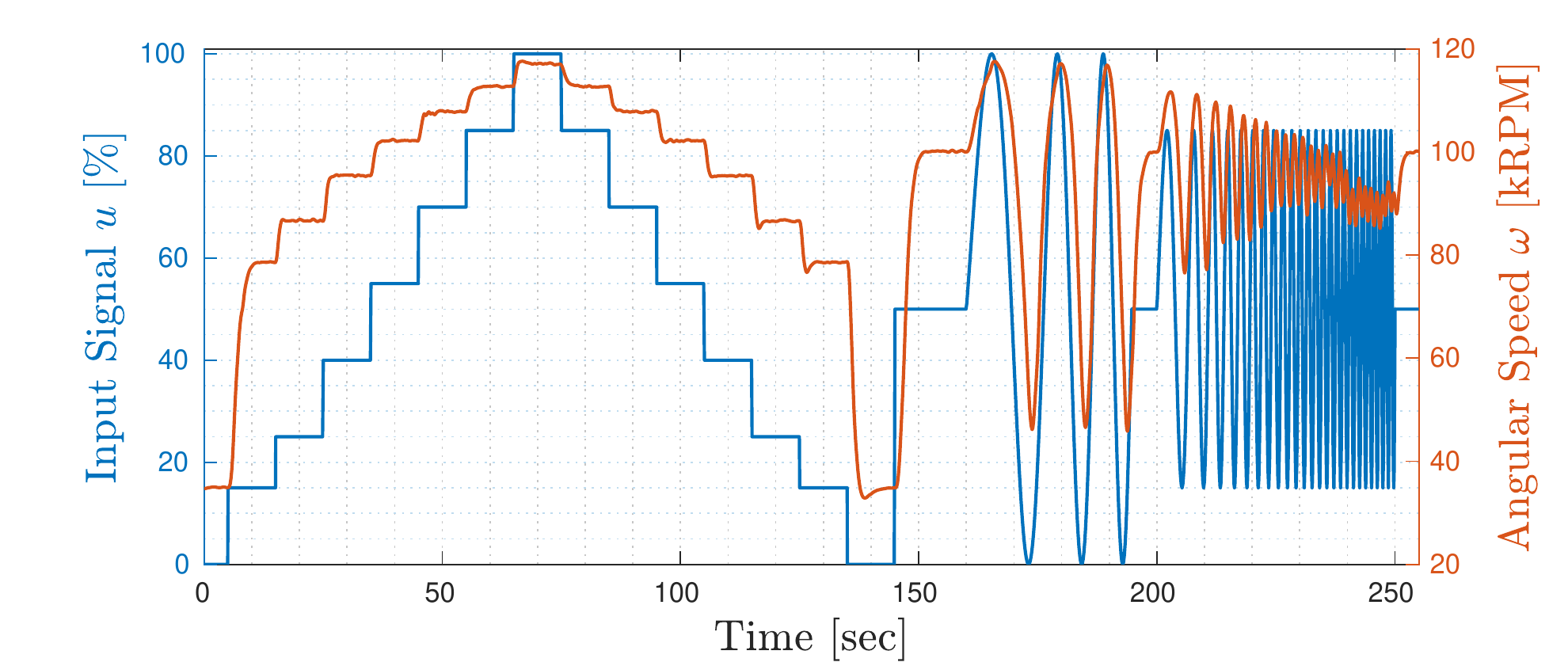}
 \caption{Steps, sinusoid, and chirp Input and Angular Speed signals}
 	\label{fig:5}
\end{figure}

The objective is to identify a dynamic model with state-space representation that describes the relationship between angular speed and input signal of the turbojet. As shown in Fig.~\ref{fig:5}, the turbojet was excited with steps, staircases, sinusoids, chirps (0.05 Hz to 0.5 Hz), and slow ramp signals in the test bench to collect data for system identification.

From a qualitative analysis of Fig.~\ref{fig:5}, it is concluded that a second-order model is an acceptable candidate. Therefore, the model to be identified is:
\begin{align}
    \ddot{\omega} = f(\omega, \dot{\omega}, u) \label{eq:7},
\end{align}
where $\omega$, $\dot{\omega}$, and $\ddot{\omega}$ are expressed in units of $\mathrm{kRPM}$, $\mathrm{kRPM/s}$, and $\mathrm{kRPM/s^2}$ respectively.

Since the measured angular speed signal is quantized in steps of 100 RPM, spline interpolation was carried out for smoothing and calculating its first and second numerical derivatives. For the thrust measurements, the Savitzky-Golay filter was used to reduce the noise and calculate the derivative.

We use the SINDy~\cite{Brunton3932} tool for model identification. SINDy is a data-driven sparse identification technique that identifies a subset of $m_{lib}$ functions from a given library of $n_{lib}$ functions using a sequentially thresholded least-squares regression in order to describe the relationship between input variables and system dynamics. The library of functions passed to SINDy can include polynomial combinations of variables or any other candidate functions. For example, for the $\omega-u$ dynamic model described in equation (\ref{eq:7}), it is possible to use the following function library up to any arbitrary degree of polynomial:
\begin{align}
    \mathbf{A} = \{1, \omega, \dot{\omega}, u, \omega^2, \dot{\omega}^2, u^2, \omega \dot{\omega}, \omega u, \dot{\omega} u, ... \} \label{set1}.
\end{align}

The SINDy algorithm then selects the functions from the given library and computes their coefficients based on the identification dataset. However, being a data-driven technique, SINDy has the following issues:
\begin{enumerate}
    \item The resulting model is derived only from the identification dataset. So, the model obtained is heavily influenced by the training dataset given to SINDy.
    \item If only polynomial terms are chosen for the library, then no intrinsic information about the system is exploited, and any relationships or constraints that describe system characteristics are overlooked.
\end{enumerate}

One way to overcome the above limitations is to incorporate known system characteristics in the candidate functions. As discussed in Sec.~\ref{sec:background}, the internal control architecture of the turbojet maps the input signal $u$ to a desired angular speed $\omega_{des}$, which is then achieved by an internal feedback control loop. Thus, it is worthwhile to investigate the steady-state characteristics between $\omega$ and $u$, because this is defined in the turbojet's ECU firmware.
\begin{figure}[t]
    \centering
    \includegraphics[scale=0.40]{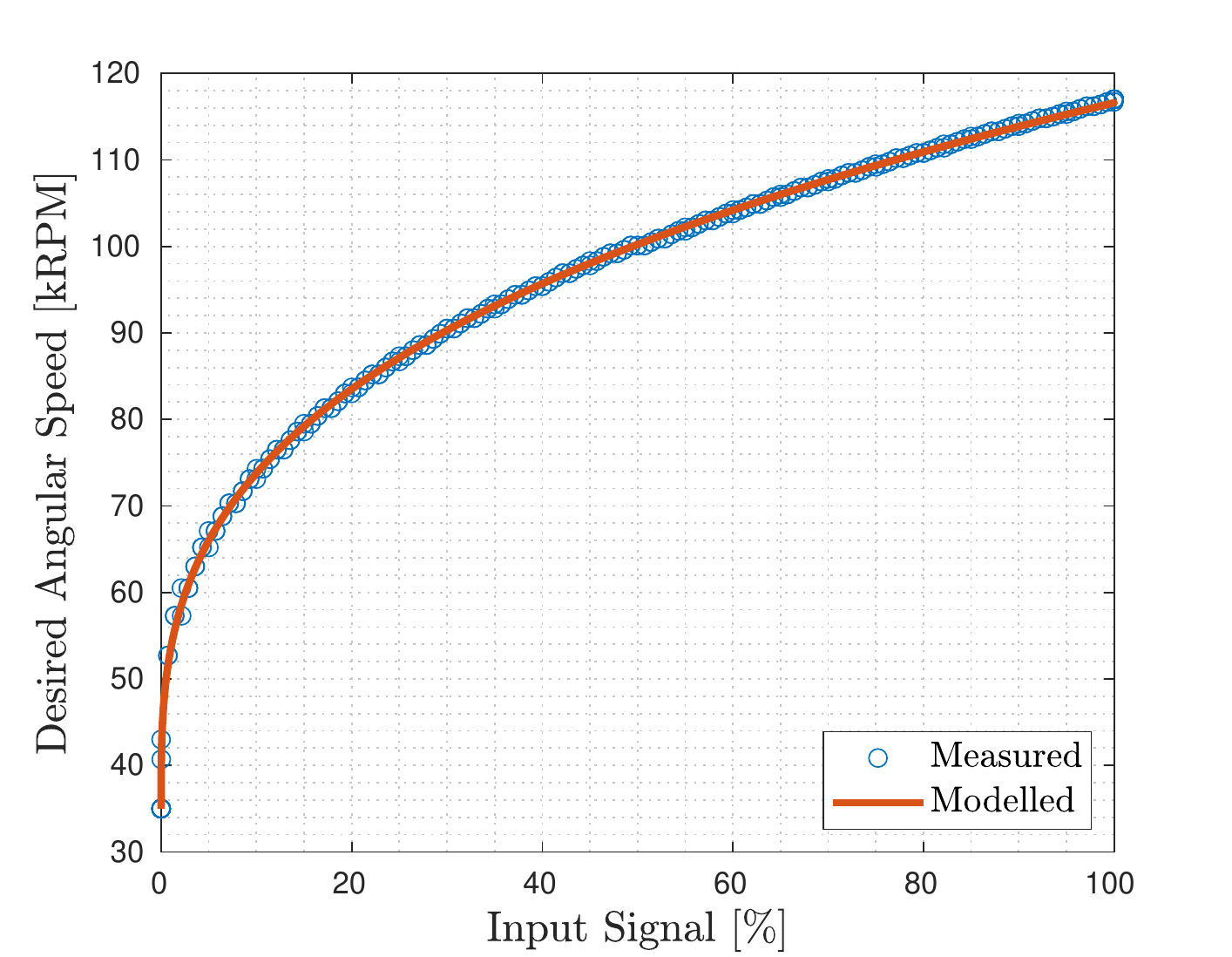}
    \caption{Measured and Modelled angular speed VS input signal at steady-state.\looseness=-1}
    \label{fig:7}
\end{figure}

From Fig.~\ref{fig:7}, we see a nonlinear monotonic relationship between $\omega$ and $u$ at steady state. We would like to model this relationship with a function $f_{ss}(\omega,u) = 0$. Note that $f_{ss}(\cdot)$ is a steady-state expression of the function $f(\cdot)$ given in Eq.~\eqref{eq:7}. We use the function in Eq.~\eqref{eqn:8} to fit the data shown in Fig.~\ref{fig:7}:
\begin{align}
    f_{ss}(\omega,u) = \omega - a_1u^b_1 - c_1 = 0 \label{eqn:8}.
\end{align}
The constants $a_1$ and $b_1$ are identified from regression on steady-state data as shown in Fig.~\ref{fig:7}, whereas $c_1$ is the idle angular speed of the turbojet in kRPM, i.e., the angular speed when $u=0$. This model fits with a 99.92\% R-square value.

So, instead of passing polynomial combinations of $\omega$, $\dot{\omega}$, and $u$ as candidate terms to SINDy (see Eq.~\eqref{set1}), we will pass the function $f_{ss}$, and all polynomial combinations of terms that necessarily contain $\dot{\omega}$. The candidate functions library will be as follows:
\begin{align}
\resizebox{.88 \columnwidth}{!}{$ 
        \mathbf{B} = \{f_{ss}(\omega,u), \dot{\omega}, \omega\dot{\omega}, u\dot{\omega}, \dot{\omega}^2, \omega^2 \dot{\omega}, u^2 \dot{\omega}, u \omega \dot{\omega}, \dot{\omega}^3,...\}
    $}.
     \label{set2}
\end{align}

\begin{figure}[t]
    \centering
    \includegraphics[scale=0.43]{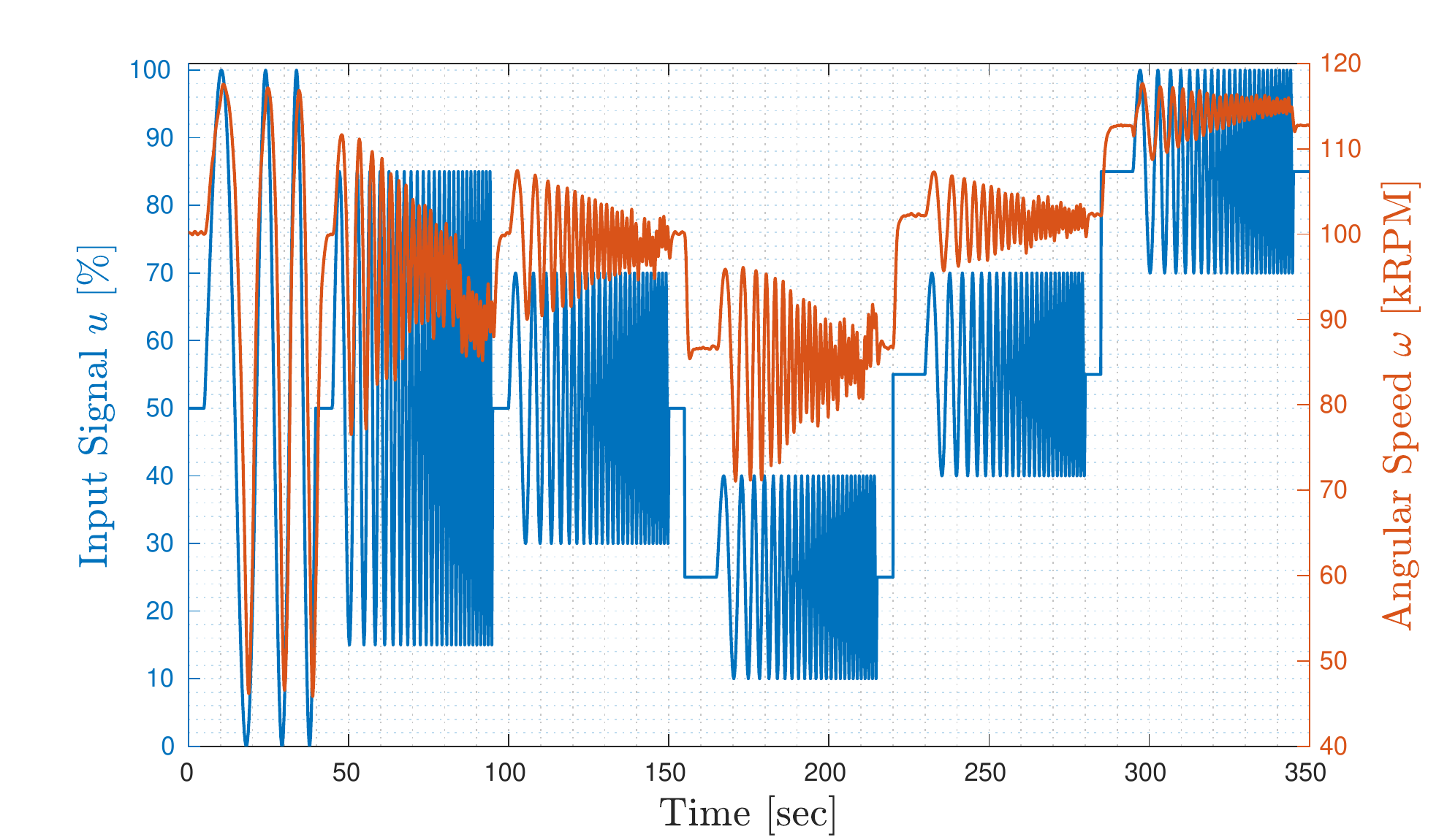}
    \caption{Identification dataset for the P220 turbojet.}
    \label{fig:8}
\end{figure}

Since the steady-state characteristics are already defined in the function $f_{ss}$, we can use fast dynamic signals (like sinusoids, and chirps) in the model identification dataset. The steady-state behaviour will always be preserved in the dynamic second-order model identified by SINDy as long as the threshold is carefully tuned. The identification dataset for SINDy is shown in Fig.~\ref{fig:8} for the P220 turbojet. A similar dataset was used for identifying the P160.

The model structure identified by SINDy is as follows:
\begin{multline}
    \ddot{\omega} = f(\omega,\dot{\omega},u)= K_{ss}(\omega - a_1u^{b_1} - c_1) + K_d \dot{\omega} +\\
    + K_{wd} \omega \dot{\omega} + K_{wwd}\omega^2 \dot{\omega} \label{eqn:11}
\end{multline}
 We also obtain the coefficients (or model parameters) $K_{ss}$, $K_d$, $K_{wd}$, and $K_{wwd}$ from SINDy. However, these can be further fine-tuned with parameter estimation EKF as discussed in the following section.

\subsection{Model parameters estimation with EKF}

After obtaining the nonlinear dynamic model in Eq.~\eqref{eqn:11} for angular speed and input signal of the turbojet from SINDy, we fine-tune the model parameters or coefficients by using the Extended Kalman Filter algorithm with parameter estimation on the identification dataset.

The augmented state vector of the dynamic state-space system is:
\begin{align} 
    \mathbf{x}_k^* = \begin{bmatrix} \omega_k & \dot{\omega}_k & K_{ss,k} & K_{d,k} & K_{wd,k} & K_{wwd,k} \end{bmatrix}^\top.
\end{align}
The state propagation to the next time step is obtained by discretizing the model in Eq.~\eqref{eqn:11}:
\begin{gather} 
    \mathbf{x}_{k+1}^* = \mathbf{x}_k^* + \begin{bmatrix} \dot{\omega}_k \\ f(\omega_k,\dot{\omega}_k,u_k) \\ [0]_{4 \times 1}                                                   \end{bmatrix} \Delta t, \\
    \mathbf{y_k} = \omega_k.
\end{gather}
The process noise covariance matrix is given by:
\begin{align}
    \mathbf{Q} = \mathrm{Diag} \begin{bmatrix} 0 & q_1 & q_2 & q_3 & q_4 & q_5 \end{bmatrix}.
\end{align}
The noise covariances $q_2$, $q_3$, $q_4$ and $q_5$ are introduced so that the EKF algorithm can change the values of the parameters at each time step.

For the initial guesses of model parameters, we use the ones obtained from SINDy. Then the EKF algorithm is applied iteratively on the same identification dataset, and the values of mean absolute error and maximum error between the modelled and measured turbojet angular speeds are recorded after each iteration. We keep iterating as long as these error values decrease with each iteration. 

The model parameters for the P160 and P220 are summarized in Table \ref{table:2}. We observe that the model parameters for the two turbojet engines are in the same numerical range. This implies that the models are coherent.

\begin{table}[t]
\caption{$\omega - u$ model parameters.}
\label{table:2}
    \centering
    \begin{tabular}{c|c|c}
    \toprule
    \textbf{Model Parameter} & \textbf{JetCat P160} & \textbf{JetCat P220}  \\
    \midrule
    $a_1$ & 19.36 & 17.68 \\
    $b_1$ & 0.3338 & 0.3332 \\
    $c_1$ & 33 & 35 \\
    $K_{ss}$ & -3.4037 & -4.4632 \\
    $K_w$ & -8.2504 & -14.5496 \\
    $K_{wd}$ & 0.1365 & 0.2883 \\
    $K_{wwd}$ & -0.0007 & -0.00165 \\
    \bottomrule
    \end{tabular}
\end{table}

\begin{figure*}[t]
    \centering
    \includegraphics[scale=0.50]{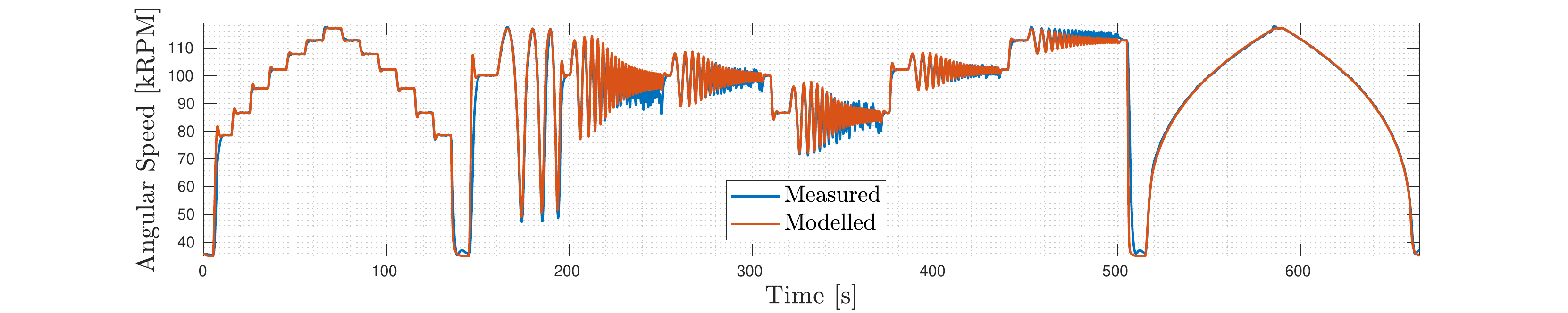} 
    \caption{Validation results for angular speed model for P220.}
    \label{fig:9}
\end{figure*}

The validation data and results of the models are presented in Fig.~\ref{fig:9} and Table \ref{table:3}, respectively. Note that the percentage errors are reported as fractions of the maximum available range of turbojet angular speed, i.e., $\omega_{max}-\omega_{idle}$.

\begin{table}[t]
\caption{$\omega - u$ model validation.}
\label{table:3}
    \centering
    \begin{tabular}{c|c|c|c}
    \toprule
    & JetCat P160 & JetCat P220 & Units \\
    \midrule
    Mean Absolute  & 1651 & 1448 & [RPM] \\
    Error & (1.8\%) & (1.8\%) & - \\
    \midrule
    Maximum Error & 44167 & 49730 & [RPM] \\
    & (49\%) & (60\%) & - \\
    \bottomrule
    \end{tabular}
\end{table}

\subsection{Angular speed - Thrust Model}

We have derived a dynamic model to predict the turbojet angular speed $\omega$ from the applied input signal $u$. But since the ultimate objective is to estimate the turbojet thrust $T$, the $T$ - $\omega$ relationship has to be investigated.

\begin{figure}[t]
    \centering
    \includegraphics[scale=0.40]{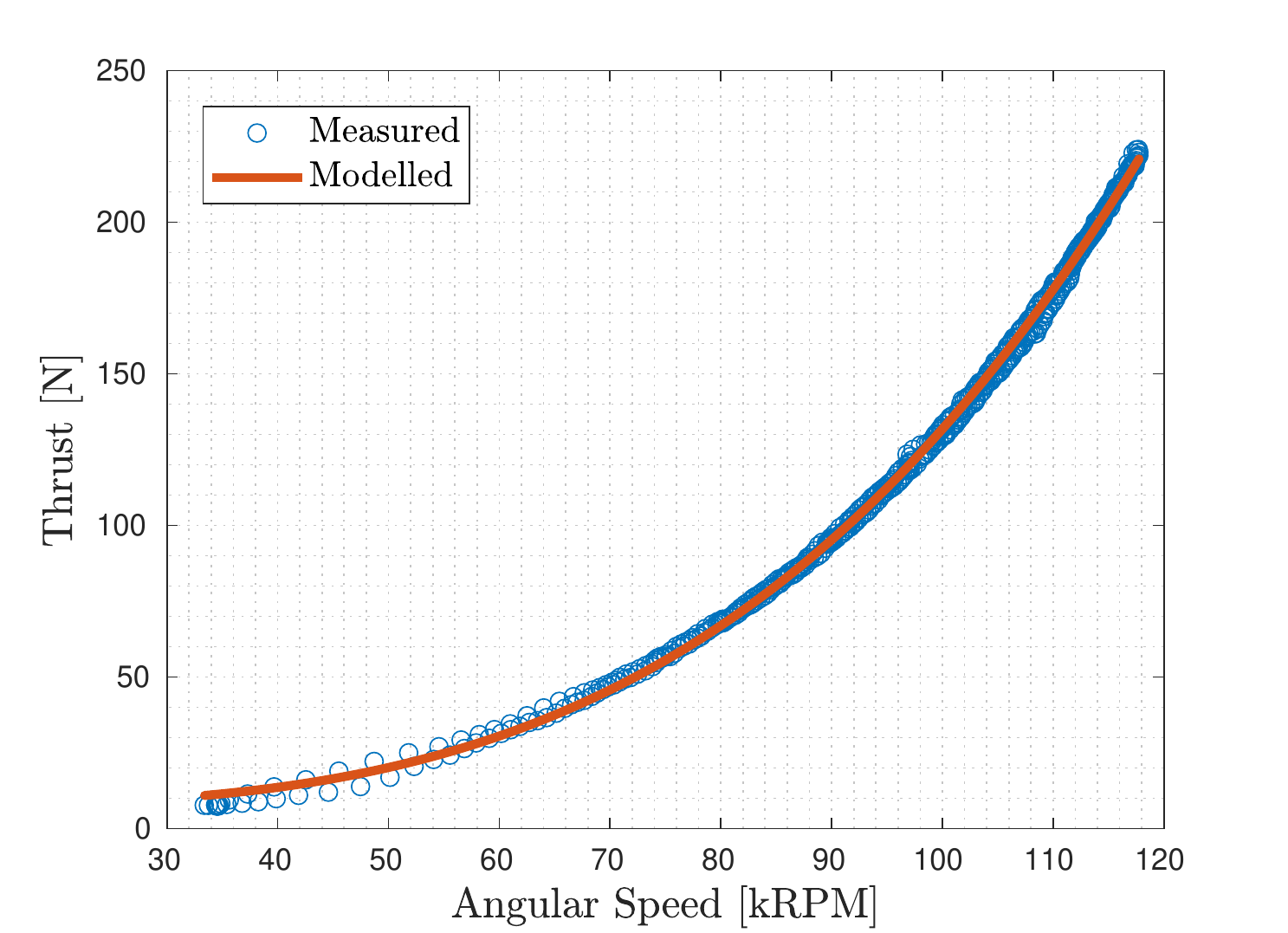}
    \caption{Thrust VS Angular speed model for P220.}
    \label{fig:10}
\end{figure}

It was discussed before in Sec.~\ref{sec:background} that the turbojet thrust depends on its angular speed, fuel flow rate and the air density. However, for subsonic turbojet with low expansion ratio nozzles, the angular speed has the most influence on thrust. Therefore, it is expected that a model relating thrust with angular speed would give a reasonably accurate prediction of thrust. This was also concluded by Jiali and Jihong in their study~\cite{Jiali2015}. The $T$ - $\omega$ model is expressed by an empirical relationship as follows:
\begin{align}
    T(\omega) = a_2 \omega^{b_2} + c_2 \label{eqn:16},
\end{align}
where $T$ is in $\mathrm{N}$ and $\omega$ is in $\mathrm{kRPM}$. The model coefficients are obtained by performing regression on experimental data as shown in Fig.~\ref{fig:10}. The coefficient values and RMSE of the $T-\omega$ model are given in Table \ref{tab:thrust-vs-angular-speed}.

\begin{table}[t]
\caption{$T - \omega$ model coefficients and RMSE.}
\label{tab:thrust-vs-angular-speed}
    \centering
    \begin{tabular}{c|c|c}
    \toprule
    & \textbf{JetCat P160} & \textbf{JetCat P220} \\
    \midrule
    $a_2$  & $4.531 \times 10^{-5}$ & $4.928 \times 10^{-5}$ \\ 
    $b_2$ & 3.136 & 3.205 \\
    $c_2$ & 4.641 & 5.477 \\
    RMSE & 1.20 N & 2.05 N \\
    \bottomrule
    \end{tabular}
\end{table}

\begin{figure}[t]
    \centering
    \includegraphics[scale=0.23]{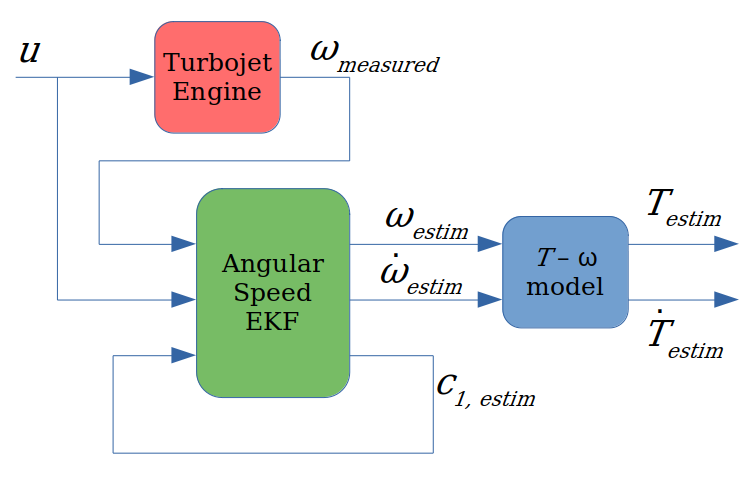}
    \caption{Extended Kalman Filter design schematic.}
    \label{fig:11}
\end{figure}
\section{OBSERVER DESIGN FOR THRUST ESTIMATION}
\label{sec:ekf_thrust}

In this section, an Extended Kalman Filter is designed to estimate the online turbojet thrust by using:
\begin{itemize}
    \item the angular speed measurements,
    \item the control input $u$,
    \item the $\omega-u$ dynamic model, and
    \item the $T-\omega$ model
\end{itemize}

\subsection{EKF Design and tuning}

In a nutshell, the EKF takes the online angular speed measurements (which are quantized), smooths them by using the $\omega-u$ dynamic process model, and then employs the $T-\omega$ model to compute the thrust and thrust derivative.

One obstacle to the 
implementation of the online EKF is turbine failure --~an occasional event that happens when a surge of air bubbles enters the fuel line 
causing 
the angular speed and thrust to drop rapidly. This poses an estimation problem because the $\omega-u$ dynamic model no longer holds in this scenario.

A simple solution to this problem could be to choose relatively high covariances for the process noise so that the EKF gives more weight to the angular speed measurements. This seems like an attractive choice because the angular speed measurements are accurate and reliable. However, because of their quantization, this would lead to noisy estimates especially for the derivative. Therefore, it seems that we have to compromise between smoothness and robustness of the thrust estimates. 

\subsection{EKF with parameter estimation}

\begin{figure*}[!t]
    \centering
    \includegraphics[scale=0.48]{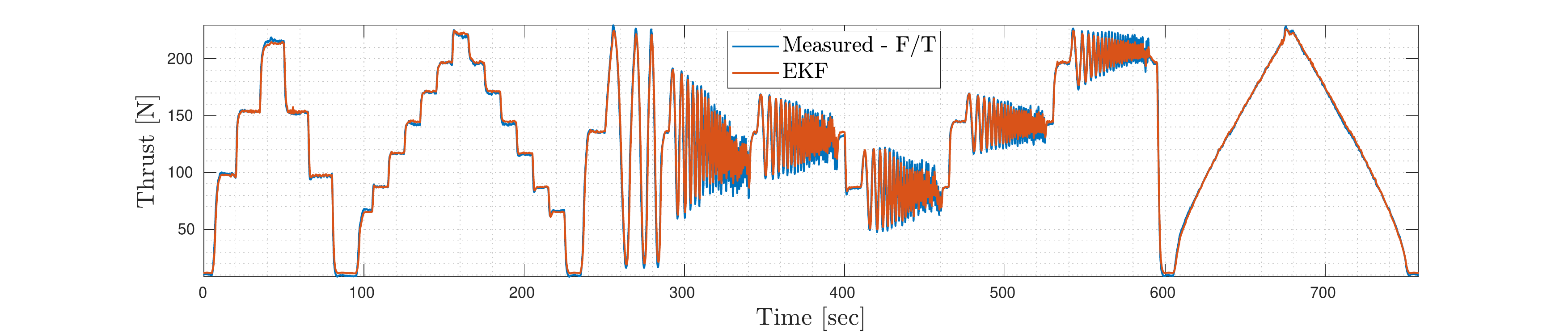} 
    \caption{Validation results for the EKF Thrust estimation for P220.}
    \label{fig:12}
\end{figure*}

\begin{figure}[!t]
    \centering
    \includegraphics[scale=0.6]{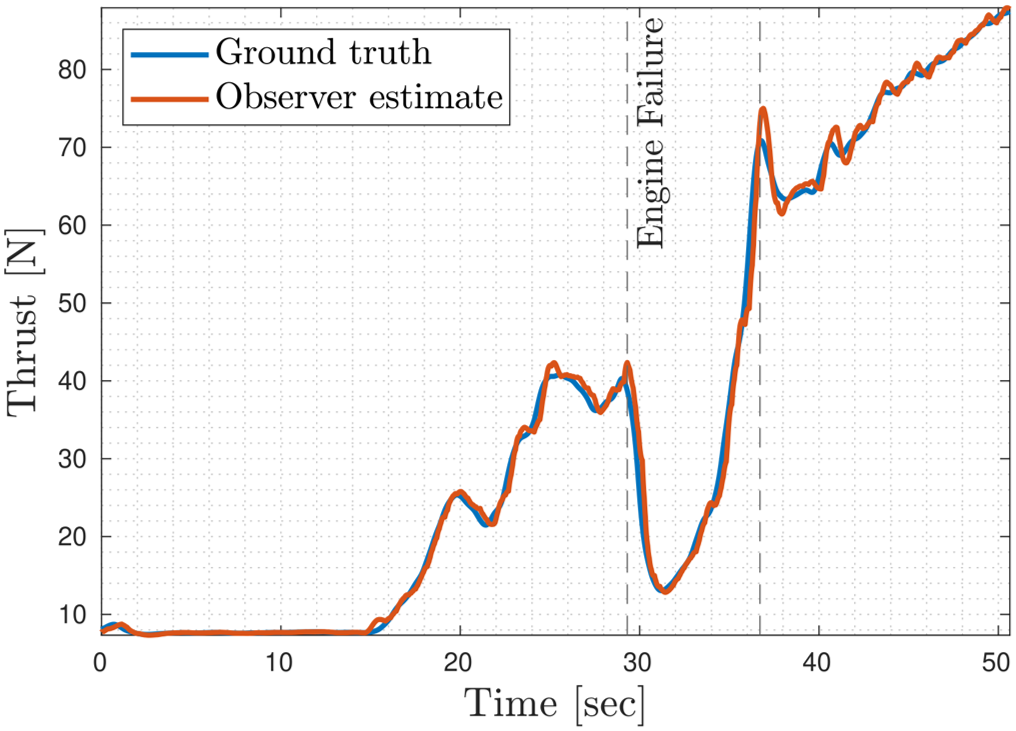} 
    \caption{EKF performance during engine failure - P160.}
    \label{fig:engine_failure}
\end{figure}

It is possible to simultaneously have both smooth and robust estimates with the parameter estimation algorithm, even during engine failures. We have the process model describing the $\omega-u$ dynamics in Eq.~\eqref{eqn:11}. The parameter $c_1$ of this dynamic model is the angular speed of the turbojet in kRPM when the input signal $u=0$, i.e, the idle angular speed. Instead of keeping this parameter constant, we can pass it to the EKF as a state to be estimated online. Thus, whenever a turbine failure happens and the angular speed drops suddenly, the EKF observes this and reduces the value of $c_1$. In this way, the estimates of angular speed are smooth and accurate in any situation. A schematic representation of the EKF is shown in Fig. \ref{fig:11}. The state and measurement vectors for this EKF are defined as follows:

\begin{align} 
    \mathbf{x}_k & = \begin{bmatrix} \omega_k & \dot{\omega}_k & c_{1,k} \end{bmatrix}^\top, \\
    \mathbf{y}_k & = \begin{bmatrix} \omega_k. \end{bmatrix}
\end{align}
The state transition is given by:
\begin{align} 
    \mathbf{x}_{k+1} & = \mathbf{x}_k + \begin{bmatrix} \dot{\omega}_k \\ f(\omega_k, \dot{\omega}_k,u_k;c_{1,k}) \\ -K_{c1} (c_{1,k}-c_1) \end{bmatrix}\Delta t,
\end{align}
where $K_{c1}$ is a non-negative constant, $c_{1,k}$ is the online value of the model parameter $c_1$ at the $k$-th time step, and $c_1$ is the original idle angular speed of the turbojet in kRPM. Notice that we have imposed first-order dynamics on $c_1$ in order to limit its variation by tuning the parameter $K_{c1}$.

The process and measurement noise covariance matrices are tuned with the help of experimental data.
\section{OBSERVER VALIDATION}
\label{sec:ekf_validation}

The EKF designed in the previous section was tuned and deployed on a test bench experiment in order to test its performance. We have the direct thrust measurements that are streamed from the F/T sensor in the test bench and filtered by the Savitzky-Golay filter. The estimated thrust from the EKF is then compared with the measured and filtered thrust from the F/T sensor to evaluate the performance and accuracy of the EKF.

Fig.~\ref{fig:12} shows the comparison between the thrust estimated by the EKF and the one measured by the F/T sensor for the P220 turbojet. A similar experiment was also run with the P160 turbojet. The error between EKF estimated thrust and measured thrust are summarized in Table \ref{tab:validation-ekf}. The percentage error is calculated as a fraction of the maximum thrust attainable by the turbojet (see Table \ref{tab:turbines_specs}). We observe that the thrust estimates provided by the EKF are in good agreement with the F/T measurements.

In Fig.~\ref{fig:engine_failure}, the EKF thrust estimates have been compared against the measured ground truth thrust during an engine failure event for the P160 turbojet. The mean absolute error in estimated thrust was found to be $1.78$~N and the max absolute error was $8.6$~N. From this, we conclude that the proposed EKF generates accurate estimates even during engine failures.

\begin{table}[t]
\caption{EKF performance indicators.}
\label{tab:validation-ekf}
    \centering
    \begin{tabular}{c|c|c}
    \toprule
    \textbf{Thrust Error} & \textbf{JetCat P160} & \textbf{JetCat P220} \\
    \midrule
    Mean Absolute  & 2.52 N & 3.96 N \\ 
    Error & (1.7\%) & (1.9\%) \\
    \midrule
    Maximum Error & 22.03 N & 42.88 N \\
    & (13.94\%) & (19.49\%) \\
    \bottomrule
    \end{tabular}
\end{table}
\section{RECAP AND CONCLUSIONS}
\label{sec:conclusions}

We started by emphasising the importance of accurate and reliable thrust estimation for jet-powered VTOL drones. The iRonCub - an aerial humanoid robot that uses four turbojet engines - is a relevant example that served as the context of this study. Limitations in engine state measurements of the turbojets used on the iRonCub posed an obstacle to accurate thrust estimation using classical techniques. So, we attempted to solve this problem with a two-step approach.

First, we used a grey-box method to identify the nonlinear system dynamics of the given small-scale turbojet engines. From the data and insights collected on our custom turbojet test bench, we built a second-order nonlinear state-space model that related the angular speed to the applied input signal. The mean absolute error of the model obtained was found to be 1.8\% on the validation dataset. We also identified a static nonlinear model that related the turbojet thrust to its angular speed.

Second, we designed an EKF that used the nonlinear model and angular speed measurements to estimate the online thrust. We exploited the parameter estimation algorithm to ensure that the EKF generated smooth and accurate estimates even in cases of turbine failure. It was observed that even for fast dynamic turbojet control signals, the EKF estimated the thrust with a high level of accuracy. The mean absolute error in thrust estimation was found to be less than 2\%, even for scenarios involving engine failure.

Though this study was done in context of the turbojet engines used on the iRonCub aerial robot, the methodology presented here can be extended to other kinds of jet engines and similar nonlinear dynamic systems. Future work would involve investigating the effects of ambient conditions and free stream inlet velocity on the given turbojet thrust. Moreover, high fidelity models that use artificial neural networks (ANNs) will be explored in order to narrow the gap between simulations and reality.

\addtolength{\textheight}{-9cm}   


\bibliographystyle{IEEEtran}
\bibliography{IEEEabrv,Biblio}

\end{document}